\pdfoutput=1

\documentclass[11pt]{article}

\usepackage[]{ACL2023}

\usepackage{times}
\usepackage{latexsym}

\usepackage{amsmath}
\usepackage{amssymb}

\usepackage{multirow}
\usepackage{multicol}
\usepackage{booktabs}
\usepackage{array}
\usepackage {graphicx}
\graphicspath{{./image/}}
\usepackage{float}
\usepackage{tablefootnote}

\usepackage[T1]{fontenc}
\usepackage{CJKutf8}
\usepackage{color}
\usepackage{bm}

\usepackage[utf8]{inputenc}

\usepackage{microtype}

\usepackage{inconsolata}

\usepackage{tikz}
\newcommand*{\circled}[1]{\lower.7ex\hbox{\tikz\draw (0pt, 0pt)%
    circle (.5em) node {\makebox[1em][c]{\small #1}};}}

\title{Block the Label and Noise: An N-Gram Masked Speller for Chinese Spell Checking}

\author{Haiyun Yang \\
  Independent Researcher \\
  \texttt{yanghy57@mail2.sysu.edu.cn} \\}

\begin{document}
\maketitle
\begin{abstract}
Recently, Chinese Spell Checking(CSC), a task to detect erroneous characters in a sentence and correct them, has attracted extensive interest because of its wide applications in various NLP tasks. Most of the existing methods have utilized BERT to extract semantic information for CSC task. However, these methods directly take sentences with only a few errors as inputs, where the correct characters may leak answers to the model and dampen its ability to capture distant context; while the erroneous characters may disturb the semantic encoding process and result in poor representations. Based on such observations, this paper proposes an n-gram masking layer that masks current and/or surrounding tokens to avoid label leakage and error disturbance. Moreover, considering that the mask strategy may ignore multi-modal information indicated by errors, a novel dot-product gating mechanism is proposed to integrate the phonological and morphological information with semantic representation. Extensive experiments on SIGHAN datasets have demonstrated that the pluggable n-gram masking mechanism can improve the performance of prevalent CSC models and the proposed methods in this paper outperform multiple powerful state-of-the-art models.
\end{abstract}

\section{Introduction}

\begin{table}
\fontsize{8pt}{11pt}\selectfont
\centering
\begin{tabular}{|l|l|}
\hline
Input & \begin{CJK*}{UTF8}{gbsn}…去外国可以认识很多的人，就可以\textcolor{red}{借(bor-}\end{CJK*}\\
&\begin{CJK*}{UTF8}{gbsn}\textcolor{red}{row)少(less)}。\end{CJK*}\\
\hline
     BERT & \begin{CJK*}{UTF8}{gbsn}
     …去外国可以认识很多的人，就可以\textcolor{red}{借(bor-}\end{CJK*}\\
     &\begin{CJK*}{UTF8}{gbsn}\textcolor{red}{row}\textcolor{blue}{绍(mediate)}。
     \end{CJK*}\\
     \hline
     Input&\begin{CJK*}{UTF8}{gbsn}…去外国可以认识很多的人，就可以\textcolor{red}{借(bor-}\end{CJK*}\\
     &\begin{CJK*}{UTF8}{gbsn}\textcolor{red}{row)}\textcolor{blue}{绍(mediate)}。\end{CJK*}\\
     \hline
     BERT&\begin{CJK*}{UTF8}{gbsn}…去外国可以认识很多的人，就可以\textcolor{blue}{介绍(in-}\end{CJK*}\\
     &\begin{CJK*}{UTF8}{gbsn}\textcolor{blue}{troduce)}。\end{CJK*}\\
    \hline
    Trans&…(Then) they can make a lot of friends in foreign\\
    &countries and introduce (themselves to foreign-\\
    &ers).\\
     \hline
\end{tabular}
\caption{An example of error disturbance. BERT  fails to correct the error \begin{CJK*}{UTF8}{gbsn}“借”\end{CJK*} if the next character is also an error \begin{CJK*}{UTF8}{gbsn}“少”\end{CJK*}. However, once  \begin{CJK*}{UTF8}{gbsn}“少”\end{CJK*} on the next position has been replaced with the correct character \begin{CJK*}{UTF8}{gbsn}“绍”\end{CJK*}, BERT  can derive the correct answer \begin{CJK*}{UTF8}{gbsn}“介”\end{CJK*}. This is a hint that the existence of error may disturb the semantic information extraction process on positions close to it. \textcolor{blue}{Correct token}/\textcolor{red}{incorrect token} is marked in \textcolor{blue}{blue}/\textcolor{red}{red}}
\vspace{-1em}
\label{error disturbance}
\end{table}

Chinese Spell Checking(CSC) requires detection and correction of erroneous characters in a Chinese sentence. Unlike English, of which the word delimiters are clearly denoted as space and it is straightforward to identify a misspelled word even without any contextual information, the basic unit for Chinese Spell Checking is usually a character in a "continuous" sentence that does not have any delimiters for words. Obviously, each character is in-vocabulary, then only in a co-occurrence with the other characters in the context, can we locate the errors and correct them.
Therefore, contextual knowledge should play a crucial role in CSC. 
Actually, nearly all of the state-of-the-art CSC models in recent years have utilized BERT\citep{devlin2019bert} to extract contextual information. HeadFilt \citep{nguyen2020adaptable}, PHMOSpell \citep{huang2021PHMOSpell} and MDCSPell \citep{zhu-etal-2022-mdcspell} directly take sentences containing a few errors as inputs for training.  GAD \citep{guo2021global} and DCN \citep{wang2021dynamic} first continue to pretrain BERT or its variant RoBERTa\citep{liu2019roberta} with a modified Masked LM task where the model is asked to predict the target from a character from confusion set, and then in the finetuning stage, the original erroneous texts are fed to a correction module to perform correction. Whichever is the case, a sentence with a few or even no errors is taken as input to produce an embedding layer for BERT  (or RoBERTa), and finally 
the whole CSC model is required to predict the original input character for each correct position  and a correct character for each erroneous position. 

Two problems have been overlooked in this process. The first problem is label leakage. Label leakage happens when there exists so strong an indicator of targets in the training input features that the model fails to generalize well in prediction where the indicator is not available. Label leakage in CSC results from both identity projection from input to target on correct positions and imbalanced distribution of correct and incorrect characters. For a correct input character, input is target as well. A simple way for BERT  model to learn the answer is to "copy" it directly through skip connection, or to focus solely on the current token through self-attention. In either case, capturing contextual information is not necessary for the model to predict the target.  And since most of the characters in the input are correct, the BERT model is easily  trained to be no longer sensitive to context, especially context in a long distance. In contrast to normal label leakage, correct input characters are available in prediction but with poor ability to capture context, such models still display unpleasant performance in prediction stage. The other problem is disturbance. For an erroneous character to be corrected in the input sentence, the semantic information extraction process of the current or adjacent positions would be disturbed by the error. Table \ref{error disturbance} presents an example of
error disturbance.

To address these issues, this paper proposes a novel n-gram masking to mask the current character and/or characters around it. More specifically, a single transformer layer is inserted between BERT embedding layer and BERT encoder. This additional layer has the same structure  as a normal transformer encoder layer does except that the query and attention mask used in multi-head attention are different. In this layer, the sum of word embeddings of the $\rm{[mask]}$ token, positional embeddings and token-type embeddings is linearly projected into query after a layernorm and dropout operation. And an n-gram attention mask where current token and/or its neighbors are assigned a large negative value is applied to the computed attention scores. All other operations remain the same as in a normal transformer  encoder layer. 

Moreover, since some erroneous characters may provide valuable phonological and morphological information for correction, blocking them may cause some side effects, a dot-product gating mechanism is designed to incorporate phonological and morphological knowledge into the model, where the interactive scores calculated by the dot-product of semantic and phonetic, semantic and graphic representations respectively are normalized by a sigmoid function to produce gate values for each modality. In summary, the contributions of this paper are in 3 folds:\\
$\bullet$ This paper proposes an innovative n-gram masking layer which alleviates common label leakage and error disturbance problems in CSC. It is proved effective on various base models across different benchmarks.\\
$\bullet$ A novel dot-product gating mechanism is designed to integrate phonological and morphological information with semantic information, which achieves superior performance against other fusion mechanisms without introduction of additional parameters or hyperparameter tuning efforts.\\
$\bullet$ The proposed methods in this paper outperform several powerful state-of-the-art models. \\

\section{Related Work}

As most NLP tasks, model structures used in CSC have evolved with the development of language model. \citet{liu2013hybrid}, \citet{yu2014chinese} utilized perplexity calculated by an n-gram language model to detect and correct errors in a sentence. Then after LSTM-based language models emerged, researchers (\citealp{wang2018hybrid}; \citealp{wang2019confusionset}) began to extract contextual information for CSC using a LSTM. In recent years, due to its powerful capacity to represent bidirectional context in a long distance, BERT has become mainstream solution for various NLP tasks, including CSC. The following section would focus on  BERT-based CSC models.  Specifically, it would introduce how the pretraining of BERT  has been adapted to CSC task and  how detection module or external knowledge is fused with BERT  backbone.

\subsection{BERT  in CSC}

BERT  utilized in CSC could either be directly employed to obtain a sequence of semantic encodings (\citealp{nguyen2020adaptable}; \citealp{huang2021PHMOSpell}; \citealp{zhu-etal-2022-mdcspell}), or be pretrained continuously with mask strategies adapted for CSC before being exploited as a semantic information extractor. Two ways \citep{li2021exploration}, DCN\citep{wang2021dynamic} and GAD \citep{guo2021global} made use of similar characters in confusion set for masking purpose. Confusion-set guided pretraining strategy strengthens model's resistance against confusion caused by errors. However, in the finetuning stage, whether with continuous pretraining or not, BERT  takes sentence with few errors as input. Such process may leak labels on correct positions or suffer from disturbance from erroneous positions as described in the introduction. 

Another limitation is that the pretraining depends on a specific confusion set. A CSC model pretrained using some confusion set could be dampaned on a new dataset because it is unable to cover new errors appearing in unseen data. Alternatively pretraining every time new data is encountered is unfeasible because pretraining on millions of sentences is expensive concerning the time and computational resources it requires.

\subsection{Correction Combined with Detection}

In order to circumvent the noise caused by erroneous characters, a considerable portion of previous researches attempted to fuse a detection module with the BERT  backbone. $\rm{BERT\_CRS}$ \citep{guo2021global} directly utilized the sequence output of the last layer of BERT  to perform a binary classification. Soft-Masked BERT  \citep{zhang2020spelling} and MDCSpell \citep{zhu-etal-2022-mdcspell} introduced additional architectures to implement detection. The former exploited a bidirectional GRU whereas the latter took advantage of the transformer block. Other works such as CRASpell \citep{liu-etal-2022-craspell}, CoSPA \citep{yang2022cospa} included a similar binary-classifier detection structure named copy mechanism. However, due to the imbalanced distribution of correct and incorrect characters, the models mentioned above is prone to merely learn the bias of data instead of knowledge that can really distinguish correct and incorrect positions, e.g. contextual knowledge.

\subsection{Fusion of Multi-Modalities}

Since $76\%$ of Chinese spelling errors belong to phonological similarity error and $46\%$ belong to visual similarity error \citep{liu2011visually}, phonological and morphological information are 2 other crucial knowledge sources for CSC task besides semantic information. Early works such as SpellGCN \citep{DBLP:journals/corr/abs-2004-14166} leveraged confusion set to encode visual and phonological information. However, it is using limited information since the confusion set is unable to cover all characters\citep{huang2021PHMOSpell}. As a result, researchers began to utilize a pretrained feature encoder to capture more general visual or phonological features. For instance, PHMOSpell \citep{huang2021PHMOSpell} exploited VGG19 \citep{simonyan2014very} and Tacotron2 \citep{shen2018natural} to encode glyph and pinyin features respectively while ReaLiSe \citep{xu2021read} utilized a trainable ResNet \citep{he2016deep}  to extract visual features. 

Generally, a fusion strategy would be applied after multi-modal features
have been encoded as vectors. ReaLiSe concatenated semantic, phonological and morphological representations together to produce gate values for 3 types of information. PHMOSpell employed relu activation and element-wise multiplication to integrate vector representations from different modalities. Both strategies encourage interaction between modalities and PHMOSpell achieves better performance on the whole according to our experiments. However, both frameworks utilized additional parameters to implement fusion and PHMOSpell requires extra hyperparameter tuning which is cumbersome. To circumvent this, this paper proposes an alternative dot-product gating mechanism that outperforms fusion strategies in PHMOSpell and ReaLiSe in most of the cases, while neither extra hyperparameters tuning nor introduction of new params is required.

\section{Approach}

\subsection{Problem Formulation}

Given a Chinese sentence consisting of n characters
\begin{math}
{X}=(x_1,x_2,...x_n)
\end{math}, the goal of Chinese Spell Checking is to predict a sequence
of output 
\begin{math}
Y=(y_1,y_2,...,y_n)
\end{math}
of equal length.There exists and only exists 2 sorts of characters concerning whether the input character is equal to the output character on the corresponding position.
Formally, \begin{math}X\end{math} can be split into 2 sets:
\begin{math}
Correct=\{x_i|x_i=y_i, i=1,2,...,n\}
\end{math}
and
\begin{math}
Incorrect=\{x_j|{x_j}\neq{y_j}, j=1,2,...n \}
\end{math}.
The set of indices of correct characters
\begin{math}
CrrPos=\{i|x_i\in{Correct}\}
\end{math}
is called correct positions whereas that of incorrect characters
\begin{math}
InCrrPos=\{j|x_j\in{Incorrect}\}
\end{math}
is called erroneous or incorrect positions.
Note that generally, the number of correct positions is far greater than that of erroneous positions, i.e.,
\begin{math}
{|CrrPos|}\gg{|InCrrPos|}
\end{math}.

\begin{figure*}[ht]
\centering
\includegraphics[scale=0.6]{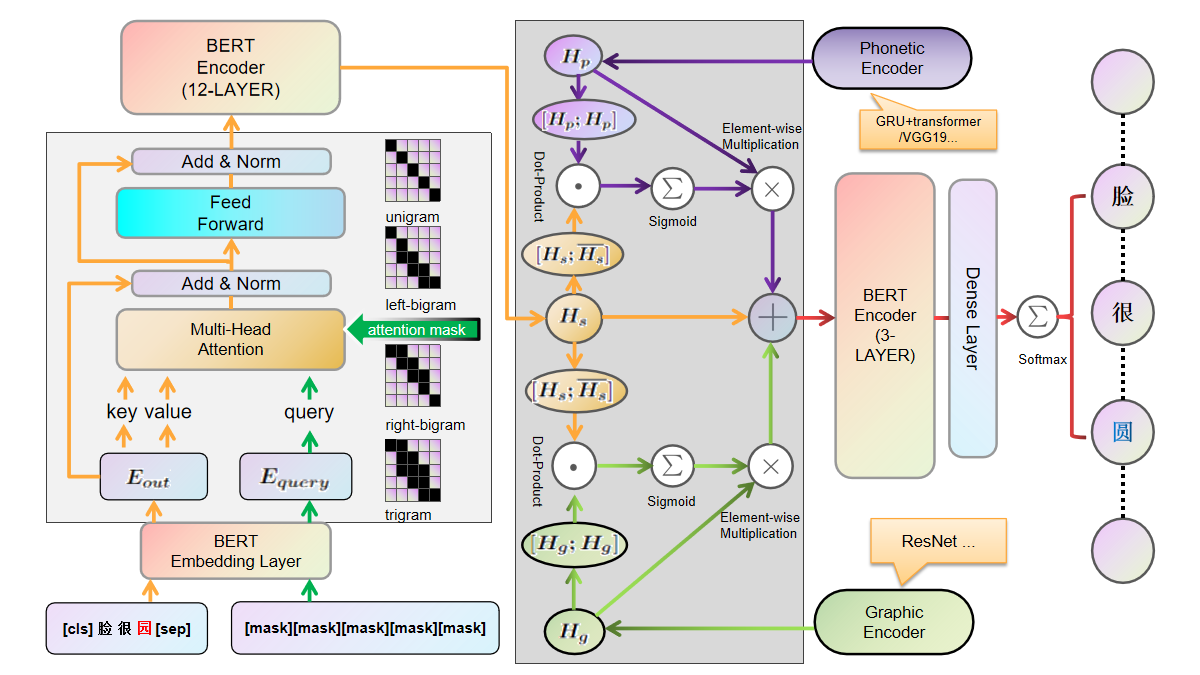}
\caption[Caption for LOF]{Overall model structure. The part in light grey is the n-gram masking layer and the part in dark grey is dot-product gating. Generally the mask positions are decided according to equation (\ref{equation13}), except that the left position of $\rm{[cls]}$ and the first character, the right position of the last character and $\rm{[sep]}$ would not be masked.\footnotemark  The input sentence "\begin{CJK*}{UTF8}{gbsn}脸很园\end{CJK*}" is supposed to mean "Face is round" but "\begin{CJK*}{UTF8}{gbsn}圆(round)\end{CJK*}" has been misspelled as "\begin{CJK*}{UTF8}{gbsn}园(garden)\end{CJK*}".}
\label{model}
\vspace{-1em} 
\end{figure*}

\subsection{Motivations}

In order to alleviate the label leakage and error disturbance, an n-gram masking strategy is proposed. As its name suggests, besides the current token, an attempt is also made to mask the surrounding tokens for there may be errors on such positions that will also disturb the current semantic extraction. Additionally this could enforce the model to pay attention to more distant context other than adjacent tokens. Moreover, in order to incorporate potential phonological and morphological knowledge in erroneous characters, which could be ignored by the n-gram masking strategy, a novel dot-product gating interaction mechanism is introduced. Technical details of this mechanism is explained in section 3.4.

In summary, the proposed model of this paper consists of 2 parts: 1) a BERT  with n-gram masking to extract semantic representations; 2) A dot-product gating strategy to fuse multi-modal information. This paper names it as D(ot-product gated)N(-gram masked)Speller. The overall architecture of DNSpeller is shown in figure \ref{model}.

\subsection{N-Gram Masking}

N-gram masking is a layer inserted between the embedding layer and 
encoder layers of a BERT. A BERT  embedding layer is defined as:
\begin{equation}
E = E_{word}+E_{pos}+E_{type}
\end{equation}
\begin{equation}
E = LayerNorm(E)
\end{equation}
\begin{equation}
E_{out} = Dropout(E)
\end{equation}
where
\begin{math}
E_{word}=(e_1^w, e_2^w, ..., e_n^w)
\end{math}
is the word embeddings of the input sequence
\begin{math}
X
\end{math}
and
\begin{math}
E_{pos}
\end{math},
\footnotetext{Left of the first and right of the last are not masked because they are correct tokens; left of $\rm{[cls]}$ and right of $\rm{[sep]}$ are not masked because there does not exist tokens on these positions.}
\begin{math}
E_{type}
\end{math}
represent corresponding positional embeddings and token type embeddings .

Then  
a single n-gram masking layer is conducted using
\begin{math}
E_{out}
\end{math}
and 
\begin{math}
E_{mask}
\end{math}
as inputs.
\begin{math}
E_{mask}
\end{math}
is the word embeddings of a sequence of $\rm{[mask]}$ tokens of equal length. An n-gram masking layer has the similar structure of a
standard BERT  encoder layer except the query hidden states and attention mask in multi-head attention are different. For simplicity, only the computation of a
single attention head is presented here:
\begin{equation}
E_{maskout} = E_{mask}+E_{pos}+E_{type}
\end{equation}
\begin{equation}
E_{maskout} = LayerNorm(E_{maskout})
\end{equation}
\begin{equation}
E_{query} = Dropout(E_{maskout})
\end{equation}
\begin{equation}
Q = E_{query}W_{Q}
\end{equation}
\begin{equation}
K = E_{out}W_{K},V = E_{out}W_{V}
\end{equation}
\begin{equation}
AttScores = \frac{QK^{T}}{\sqrt{d_k}}
\end{equation}
\begin{equation}
AttScores = AttScores+AttMask
\end{equation}
\begin{equation}
head = Softmax(AttScores)V
\end{equation}
where 
equation (4) shares the same positional embeddings and token type embeddings with equation (1) and
\begin{math}
AttMask
\end{math} is defined as follows:\\
\begin{equation}
AttMask_{i,j} = \begin{cases}
                -10000., \quad{j}\in {MaskIndices}\\
                
                0., \quad{j}\notin {MaskIndices}\\
                \end{cases}
\end{equation}
where
\begin{math}
AttMask_{i,j}
\end{math}
means the mask score of the ith token with repect to the jth token.
Different choices of n-gram lead to different
\begin{math}
{MaskIndices}
\end{math}:\\
\begin{equation}
MaskIndices = \begin{cases}
              \{i\}, \quad{unigram}\\
              \{i-1, i\}, \quad{left-bigram}\\
              \{i, i+1\}, \quad{right-bigram}\\
              \{i-1, i, i+1\}, \quad{trigram}\\
              \end{cases}
\label{equation13}
\end{equation}

The output of a single n-gram masking layer is then taken as input for the BERT encoder, and the sequence of hidden states of the last layer of BERT
\begin{math}
H_{s} = (h_{1}^{s}, h_{2}^{s},...,h_{n}^{s})
\end{math} is the extracted semantic encodings of input 
\begin{math}
X
\end{math}.

\subsection{Dot-Product Gating}

To better integrate semantic information generated by an n-gram-masked BERT  with potential knowledge in spelling errors, a dot-product gating mechanism is designed as follows:
\begin{equation}
Scores_{p} = [H_{s};\overline{H_{s}}]\cdot[H_{p};H_{p}]
\end{equation}
\begin{equation}
Gates_{p} = sigmoid(Scores_{p})
\end{equation}
\begin{equation}
Scores_{g} =
 [H_{s};\overline{H_{s}}]\cdot[H_{g};H_{g}]
\end{equation}
\begin{equation}
Gates_{g} = sigmoid(Scores_{g})
\end{equation}
where 
\begin{math}
H_{s}
\end{math},
\begin{math}
\overline{H_{s}}
\end{math},
\begin{math}
H_{p}
\end{math},
\begin{math}
H_{g}
\in{\mathbb{R}^{n\times{d}}}
\end{math}, and n, d stand for sequence lenghth and hidden size of encodings of each modality respectively.
\begin{math}
\overline{H_{s}} 
\end{math}
is the mean of 
\begin{math}
H_{s}
\end{math}
along the axis of sequence length while 
\begin{math}
H_{p}
\end{math}
and
\begin{math}
H_{g}
\end{math}
are phonological and morphological encodings of the input sequence. This paper utilized the same phonetic and graphic encoder as in ReaLiSe \citep{xu2021read} to generate these encodings, in which a GRU \citep{cho2014learning} followed by a 4-layer transformer is employed to encode phonetics and a ResNet \citep{he2016deep} is exploited to extract graphic information. The 
reason for using identical encoders is to exclude any factors that may lead to a different performance except the interactive mechanism, so that a fair comparison can be conducted between different fusion strategies.
$[\quad]$ concatenates 2 matrices along the axis of hidden size, while $\cdot$ means dot-product operation for each time step, i.e. the shape of $Scores_{p}$ and $Scores_g$
would become $n\times1$.
After computing the gate values for phonological and morphological modalities, the encodings are fused as follows:
\begin{equation}
Fusion = H_{s}+{Gates_{p}} \circled{$\times$}{H_{p}}+{Gates_{g}}\circled{$\times$}{H_{g}}
\end{equation}
where 
\begin{math}
Gates_{p}
\end{math}
and 
\begin{math}
Gates_{g}
\end{math}
are broadcasted along the dimension with a size of 1 and \textcircled{$\times$} is element-wise multiplication.
And then 
\begin{math}
Fusion
\end{math}
is inputted to a 3-layer BERT encoder:
\begin{equation}
H_{l} = Transformer(H_{l-1}), l\in{[1,3]}
\end{equation}, where
\begin{equation}
H_0 = Fusion
\end{equation}
In order to produce a probability distribution in terms of vocabulary, the final encoding output
\begin{math}
H_{3}
\end{math}
is first projected to vocab size through a dense layer and then normalized by softmax:
\begin{equation}
Probs = Softmax(H_{3}W_{o}+b_{o})
\end{equation}
\begin{math}
Probs
\end{math} can either be used to calculate cross-entropy loss in training stage or to obtain the predicted token with highest probability in prediction stage.
\begin{table*}[t]
\fontsize{8.5pt}{9.5pt}\selectfont
\renewcommand\arraystretch{1.2}
\resizebox{\linewidth}{!}{
\begin{tabular}{|m{2cm}<{\centering}|m{3cm}<{\centering}|m{4cm}<{\centering}|m{4cm}<{\centering}|}
\hline
\multirow{1}{*}[-1.5ex]{Dataset} & \multirow{1}{*}[-1.5ex]{Model} & Detection Level& Correction Level\\
& & Pre\qquad{Rec}\qquad{F1} & Pre\qquad{Rec}\qquad{F1}\\
\hline
\multirow{2}{*}[-6.5ex]{\centering{SIGHAN13}} & $\rm{BERT }$ & 87.00\quad{80.64}\quad{83.70} &85.78\quad{79.51}\quad{82.52}\\
&$\rm{NM-BERT}$ & \textcolor{blue}{\underline{88.33}}\quad{\textcolor{blue}{\underline{81.87}}}\quad{\textcolor{blue}{\underline{84.98}}} &\textcolor{blue}{\underline{87.22}}\quad{\textcolor{blue}{\underline{80.84}}}\quad{\textcolor{blue}{\underline{83.91}}}\\
\cline{2-4}
&$\rm{SpellGCN^\dag}$ & 82.19\quad{73.05}\quad{77.35} &80.50\quad{71.54}\quad{75.76}\\
&$\rm{NM-SpellGCN^{*\dag}}$ &\textcolor{blue}{\underline{84.15}}\quad{\textcolor{blue}{\underline{74.45}}}\quad{\textcolor{blue}{\underline{79.00}}} &\textcolor{blue}{\underline{82.79}}\quad{\textcolor{blue}{\underline{73.25}}}\quad{\textcolor{blue}{\underline{77.72}}}\\
\cline{2-4}
&$\rm{DCN}$ & 85.32\quad{77.45}\quad{81.20} &83.55\quad{75.85}\quad{79.52}\\
&$\rm{NM-DCN^*}$ & \textcolor{blue}{\underline{86.41}}\quad{\textcolor{blue}{\underline{78.36}}}\quad{\textcolor{blue}{\underline{82.19}}} &\textcolor{blue}{\underline{85.41}}\quad{\textcolor{blue}{\underline{77.45}}}\quad{\textcolor{blue}{\underline{81.24}}}\\
\hline
\multirow{2}{*}[-6.5ex]{\centering{SIGHAN14}} & $\rm{BERT }$ & 64.84\quad{68.08}\quad{66.42} &62.82\quad{65.96}\quad{64.35}\\
&$\rm{NM-BERT}$ & \textcolor{blue}{\underline{65.44}}\quad{\textcolor{blue}{\underline{68.46}}}\quad{\textcolor{blue}{\underline{66.92}}} &\textcolor{blue}{\underline{63.97}}\quad{\textcolor{blue}{\underline{66.92}}}\quad{\textcolor{blue}{\underline{65.41}}}\\
\cline{2-4}
&$\rm{SpellGCN}$ & 65.88\quad{67.98}\quad{66.91} &62.77\quad{64.78}\quad{63.76}\\
&$\rm{NM-SpellGCN^*}$ & 65.69\quad{67.42}\quad{66.54} &\textcolor{blue}{\underline{63.85}}\quad{\textcolor{blue}{\underline{65.54}}}\quad{\textcolor{blue}{\underline{64.68}}}\\
\cline{2-4}
&$\rm{DCN}$ & 64.65\quad{66.48}\quad{65.55} &63.74\quad{65.54}\quad{64.62}\\
&$\rm{NM-DCN^*}$&\textcolor{blue}{\underline{65.73}}\quad{66.10}\quad{\textcolor{blue}{\underline{65.92}}} &\textcolor{blue}{\underline{64.61}}\quad{64.97}\quad{\textcolor{blue}{\underline{64.79}}}\\
\hline
\multirow{2}{*}[-6.5ex]{\centering{SIGHAN15}} & $\rm{BERT }$ & 75.48\quad{80.22}\quad{77.78} &73.39\quad{78.00}\quad{75.63}\\
&$\rm{NM-BERT^* }$ & \textcolor{blue}{\underline{75.65}}\quad{\textcolor{blue}{\underline{80.41}}}\quad{\textcolor{blue}{\underline{77.96}}} &\textcolor{blue}{\underline{73.91}}\quad{\textcolor{blue}{\underline{78.56}}}\quad{\textcolor{blue}{\underline{76.16}}}\\
\cline{2-4}
&$\rm{SpellGCN}$ & 76.45\quad{79.09}\quad{77.75} &73.64\quad{76.18}\quad{74.89}\\
&$\rm{NM-SpellGCN^*}$& 76.40\quad{77.09}\quad{76.74} &\textcolor{blue}{\underline{74.60}}\quad{75.27}\quad{\textcolor{blue}{\underline{74.93}}}\\
\cline{2-4}
&$\rm{DCN}$ & 76.06\quad{78.55}\quad{77.28} &72.89\quad{75.27}\quad{74.06}\\
&$\rm{NM-DCN}$ & \textcolor{blue}{\underline{76.90}}\quad{\textcolor{blue}{\underline{79.27}}}\quad{\textcolor{blue}{\underline{78.07}}} &\textcolor{blue}{\underline{74.78}}\quad{\textcolor{blue}{\underline{77.09}}}\quad{\textcolor{blue}{\underline{75.92}}}\\
\hline
\end{tabular}}
\caption{Results on BERT, SpellGCN, and DCN across 3 SIGHAN test sets. "$\rm{NM-(Model-X})$" denotes the n-gram masked version of the corresponding base model and results which are underlined and marked in blue indicate that there's an improvement on the n-gram masked version against the base. Model name marked with "*" means n-gram masking strategy is activated only in training and needs not to be turned on in prediction stage. Models with "$\dag$" adopt the convention in the original paper of SpellGCN where  the prediction model for SIGHAN13 is further finetuned on the SIGHAN13 training set instead of removing "de" corrections directly as in the other models. F1 of correction level is the key metric which reflects the overall and final performance of a CSC system.}
\label{main results 1}
\vspace{-1em}
\end{table*}

\section{Experiments}

\subsection{Datasets}

Following most of the previous works, both manually labeled SIGHAN datasets (\citealp{wu2013chinese}; \citealp{yu2014overview}; \citealp{tseng2015introduction}) and automatically generated corpus \citep{wang2018hybrid} are utilized in the experiments, among which the concatenation of the training sets of 3 SIGHAN datasets and the automatically generated 271K samples are used for training, and 3 test sets of SIGHAN are for evaluation. The statistics of data is shown in Appendix \ref{appendix:label statistics}.

\begin{table*}[t]
\fontsize{8.5pt}{9.5pt}\selectfont
\renewcommand\arraystretch{1.2}
\resizebox{\linewidth}{!}{
\begin{tabular}{|m{2cm}<{\centering}|m{3cm}<{\centering}|m{4cm}<{\centering}|m{4cm}<{\centering}|}
\hline
\multirow{1}{*}[-1.5ex]{Dataset} & \multirow{1}{*}[-1.5ex]{Model} & Detection Level& Correction Level\\
& & Pre\qquad{Rec}\qquad{F1} & Pre\qquad{Rec}\qquad{F1}\\
\hline
\multirow{2}{*}[-6.5ex]{\centering{SIGHAN13}}&$\rm{PHMOSpell}$ & 88.38\quad{83.01}\quad{85.61} &87.39\quad{82.08}\quad{84.65}\\
&$\rm{NM-PHMOSpell}$ &88.25\quad{82.80}\quad{85.44} &86.61\quad{81.26}\quad{83.85}\\
\cline{2-4}
&$\rm{ReaLiSe}$ &87.71\quad{82.29}\quad{84.91} &86.17\quad{80.84}\quad{83.42}\\
&$\rm{NM-ReaLiSe}$ &\textcolor{blue}{\underline{87.94}}\quad{81.87}\quad{84.80} &\textcolor{blue}{\underline{86.73}}\quad{80.74}\quad{\textcolor{blue}{\underline{83.63}}}\\
\cline{2-4}
&$\rm{DGSpeller}$ &88.58\quad{\textbf{83.11}}\quad{\textbf{85.76}} &\textbf{87.71}\quad{\textbf{82.29}}\quad{\textbf{84.91}}\\
&$\rm{DNSpeller^*}$ &\textcolor{blue}{\textbf{\underline{88.78}}}\quad{82.29}\quad{85.41} &87.67\quad{81.26}\quad{84.34}\\
\hline
\multirow{2}{*}[-6.5ex]{\centering{SIGHAN14}} &$\rm{PHMOSpell}$ & 65.50\quad{68.65}\quad{67.04} &63.30\quad{66.35}\quad{64.79}\\
&$\rm{NM-PHMOSpell}$ & \textcolor{blue}{\textbf{\underline{66.12}}}\quad{\textcolor{blue}{\textbf{\underline{69.81}}}}\quad{\textcolor{blue}{\textbf{\underline{67.91}}}} &\textcolor{blue}{\underline{64.12}}\quad{\textcolor{blue}{\underline{67.69}}}\quad{\textcolor{blue}{\underline{65.86}}}\\
\cline{2-4}
&$\rm{ReaLiSE}$ & 63.99\quad{67.31}\quad{65.60} &62.34\quad{65.58}\quad{63.92}\\
&$\rm{NM-ReaLiSe^*}$ & \textcolor{blue}{\underline{65.99}}\quad{\textcolor{blue}{\underline{69.04}}}\quad{\textcolor{blue}{\underline{67.48}}} &\textcolor{blue}{\textbf{\underline{64.71}}}\quad{\textcolor{blue}{\underline{67.69}}}\quad{\textcolor{blue}{\underline{66.17}}}\\
\cline{2-4}
&$\rm{DGSpeller}$ & 66.11\quad{69.04}\quad{67.54} &64.64\quad{67.50}\quad{66.04}\\
&$\rm{DNSpeller}$ & 65.93\quad{\textcolor{blue}{\underline{69.23}}}\quad{67.54} &\textcolor{blue}{\underline{64.65}}\quad{\textcolor{blue}{\textbf{\underline{67.88}}}}\quad{\textcolor{blue}{\textbf{\underline{66.23}}}}\\
\hline
\multirow{2}{*}[-6.5ex]{\centering{SIGHAN15}}&$\rm{PHMOSpell}$ & 76.23\quad{80.04}\quad{78.09} &75.35\quad{79.11}\quad{77.19}\\
&$\rm{NM-PHMOSpell^*}$ & \textcolor{blue}{\textbf{\underline{78.31}}}\quad{\textcolor{blue}{\textbf{\underline{82.07}}}}\quad{\textcolor{blue}{\textbf{\underline{80.14}}}} &\textcolor{blue}{\textbf{\underline{77.25}}}\quad{\textcolor{blue}{\textbf{\underline{80.96}}}}\quad{\textcolor{blue}{\textbf{\underline{79.06}}}}\\
\cline{2-4}
&$\rm{ReaLiSe}$ & 75.97\quad{79.48}\quad{77.69} &74.56\quad{78.00}\quad{76.24}\\
&$\rm{NM-ReaLiSe^*}$ &\textcolor{blue}{\underline{77.82}}\quad{\textcolor{blue}{\underline{81.70}}}\quad{\textcolor{blue}{\underline{79.71}}} &\textcolor{blue}{\underline{76.76}}\quad{\textcolor{blue}{\underline{80.59}}}\quad{\textcolor{blue}{\underline{78.63}}}\\
\cline{2-4}
&$\rm{DGSpeller}$ &76.13\quad{80.78}\quad{78.39} &74.91\quad{79.48}\quad{77.13}\\
&$\rm{DNSpeller}$ &\textcolor{blue}{\underline{78.04}}\quad{80.78}\quad{\textcolor{blue}{\underline{79.38}}} &\textcolor{blue}{\underline{76.96}}\quad{\textcolor{blue}{\underline{79.67}}}\quad{\textcolor{blue}{\underline{78.29}}}\\
\hline
\end{tabular}}
\caption{Main results on PHOMSpell, ReaLiSe and DNSpeller across 3 SIGHAN datasets. Notations are the same as tabel \ref{main results 1} except that DGSpeller is a multi-modal speller that fuses the features with dot-product gating, and DNSpeller is the n-gram masked version of DGSpeller. The best results are marked in bold.}
\label{main results 2}
\end{table*}
\begin{table}[t]
    \centering
    \fontsize{9pt}{15pt}\selectfont
    \begin{tabular}{c|c|c|c}
    \hline
    \multirow{2}{*}{N-gram}& \multicolumn{3}{c}{Correction-F1}\\ 
    \cline{2-4}
   &SIGHAN13&SIGHAN14&SIGHAN15\\
    \hline
    none & 82.52 & 64.35 & 75.63\\
    unigram & 82.48 & 62.52 & 75.43\\
    left-bigram & 82.89 & 63.64 & 75.64\\
    right-bigram & \textbf{83.91} & 64.17 & 76.03\\
    trigram & 82.20 & \textbf{65.41} & \textbf{76.16}\\
    \hline
    \end{tabular}
    \caption{Results of BERT using different n-grams. Best results are in bold.}
    \vspace{-1em} 
    \label{results of bert with different n-grams}
\end{table}
\begin{table}[ht]
    \centering
    \fontsize{8pt}{15pt}\selectfont
    \begin{tabular}{c|c|c|c}
    \hline
    Model & SIGHAN13 & SIGHAN14 & SIGHAN15\\
    \hline
    NM-BERT & 2L, \textbf{2R} & \textbf{3} & 2L, 2R, \textbf{3}\\
    \hline
    NM-SpellGCN & 2L, 2R, \textbf{3} & 2L, \textbf{2R}, 3 & \textbf{2R}\\
    \hline
    NM-ReaLiSe  & \textbf{3} & 2R, \textbf{3} & all(\textbf{2L})\\
    \hline
    NM-DCN   & 2L, 2R, \textbf{3} & \textbf{2R} & all(\textbf{2L})\\
    \hline
    NM-PHMOSpell & null &1, \textbf{2R}, 3 & all(\textbf{2L})\\
    \hline
    DNSpeller & null &\textbf{2L}&1, 2L, \textbf{3}\\
    \hline
    \end{tabular}
    \caption{n-grams that gain improvements. 2R and 2L means right-bigram and left-bigram respectively, while 1 and 3 stand for unigram and trigram. The best n-gram option is marked in bold. }
    \vspace{-1em}
    \label{n-grams that gain improvements}
\end{table}

\subsection{Baselines}

This paper uses the following models as baselines to compare with the proposed approach:

$\bullet$ ReaLiSe \citep{xu2021read} exploits ResNet and GRU with a transformer encoder to extract graphic and phonetic information, and integrates them with the semantic representations using a selective modality fusion module.

$\bullet$ PHMOSpell \citep{huang2021PHMOSpell} is also a multi-modal CSC model which applies an adaptive gating strategy to fuse semantic, phonological and morphological encodings together. But unlike ReaLiSe, it uses 2 embedding lookup tables to directly generate phonological and morphological encodings.  

$\bullet$ DCN \citep{wang2021dynamic} captures relation of adjacent candidate corrections generated by a RoBERTa combined with a pinyin enhanced generator. This model is required to consider phonetic, semantic and adjacent connection simultaneously to produce the final corrections.

$\bullet$ SpellGCN \citep{DBLP:journals/corr/abs-2004-14166} incorporates phonological and shape similarities via a graph convolutional network(GCN). Character encodings outputted by GCN consist of the weights of the last dense layer, where classification over vocabulary is implemented.

$\bullet$ BERT \citep{devlin2019bert} is initialized with weights from BERT-wwm model \citep{cui2021pre}. Sequence output of the last layer is leveraged to predict target for each input token. 
\label{baselines}

\subsection{Experimental Setup and Metrics}
This paper implements strict comparisons between the baselines and their n-gram masked versions. Detailed settings are presented in Appendix \ref{appendix: experimental setup}. As in most previous works, this paper reports sentence-level metrics of detection and correction, among which correction-F1 is the vital metric because it reflects the overall and final performance of a CSC system. 

\subsection{Experimental Results}

\subsubsection{The Effectiveness of N-Gram Masking}

According to table \ref{main results 1} and table \ref{main results 2}, the improvements of correction-F1 by applying n-gram masking can be observed in most of the cases. Especially NM-PHMOSpell and DNSpeller demonstrate best correction-F1s on SIGHAN15 and SIGHAN14 respectively with the n-gram masking employed, outperforming all the baselines in section \ref{baselines} by  mean margins of 3.46\% and 1.94\%. On average, the n-gram masked model 
outperforms its base by a margin of 1.31\% on SIGHAN15, 0.94\% on SIGHAN14, and 0.65\% on SIGHAN13; whereas from the perspective of CSC model, mean improvements of 1.62\%, 1.25\%, 0.99\% of 3 SIGHAN benchmarks are observed in ReaLiSe, DCN, and BERT in comparison with their n-gram masked versions; while averaged gains of 0.97\%, 0.71\%, 0.26\% are shown in the cases of SpellGCN, PHMOSpell and DGSpeller. These results consistently indicate that the n-gram masking strategy is effective and alleviates a universal problem in BERT-based CSC models. 

Moreover, NM-BERT performs better than ReaLiSe on SIGHAN13, outperforms both ReaLiSe and PHMOSPell on SIGHAN14 and achieves a competitive performance with ReaLiSe with a very small margin of 0.08\% on SIGHAN15 in terms of correction-F1. Note that NM-BERT marked with "*" on SIGHAN15 doesn't require the n-gram masking layer to be activated during prediction, which would not cause any increment of computational complexity. All the phenomena above imply that with a very thin model structure, NM-BERT can achieve better or competitive performance compared to sophisticated methods that introduce extra networks to model external knowledge, therefore is a practical choice for CSC task in the real world.

\subsubsection{The Effectiveness of Dot-Product Gating}

Table \ref{main results 2} shows the results of 3 models that apply
different fusion strategies to multi-modal features, where DGSpeller and DNSpeller with the dot-product gating applied achieve the highest correction-F1 scores on SIGHAN13 and SIGHAN14 respectively, exceeding the baselines in section \ref{baselines} by 2.38\% and 1.94\% on average. DGSpeller outperforms ReaLiSe by 2.12\% on SIGHAN14, 1.49\% on SIGHAN13, and 0.89\% on SIGHAN15, while exceeds PHMOSpell by an averaged margin of 0.48\%. On the other hand, when combined with the n-gram masking strategy, DNSpeller still achieves better performance than NM-ReaLiSe and NM-PHMOSpell in terms of the mean correction-F1 of 3 SIGHAN benchmarks. 

It is worth noting that adaptive gating in PHMOSpell requires additional tuning on the weights of phonetic and graphic features and ReaLise introduces extra parameters to implement selective modality fusion module, while the proposed dot-product gating mechanism decides the weights of multi-modal features dynamically through simple dot-product gating without the need to tune hyperparameters or to introduce additional params.

\subsubsection{Effects of Different N-Grams}

Table \ref{results of bert with different n-grams} presents the results of different n-gram masked BERTs, where best performance is achieved using bigram masking on SIGHAN13, and trigram masking on SIGHAN14 and 15. Table \ref{n-grams that gain improvements} lists all the n-gram options that lead to a performance gain of correction-F1, most of which are bigram and trigram. This phenomenon is in line with the expectation that comparing to unigram masking, bigram and trigram masking that mask surrounding positions besides the current token can better help to filter out the noise from adjacent incorrect characters, and to drive the attention of the CSC model to more distant context. Cases that demonstrate the proposed n-gram masking strategy enables CSC models to capture more distant context are shown in table \ref{label leakage} and would be analyzed in detail in the next section.

\begin{table}[t]
    \fontsize{8pt}{11pt}\selectfont
    \centering
    \begin{tabular}{|l|l|}
     \hline
     Input &\begin{CJK*}{UTF8}{gbsn}或是有的老公很自私要求自己的老婆得在家照\end{CJK*}\\
     &\begin{CJK*}{UTF8}{gbsn}顾家庭不让\textcolor{red}{他(him)}去上班。\end{CJK*}\\
     \hline
     ReaLiSe &\begin{CJK*}{UTF8}{gbsn}或是有的老公很自私要求自己的老婆得在家照\end{CJK*}\\
     &\begin{CJK*}{UTF8}{gbsn}顾家庭不让\textcolor{red}{他(him)}去上班。\end{CJK*}\\
     \hline
     NM- &\begin{CJK*}{UTF8}{gbsn}或是有的老公很自私要求自己的老婆得在家照\end{CJK*}\\
     ReaLiSe&\begin{CJK*}{UTF8}{gbsn}顾家庭不让\textcolor{blue}{她(her)}去上班。\end{CJK*}\\
     \hline
     Trans  &Or the husband is selfish and asks his wife to stay \\
     & at home to take care of the family, and doesn't al-\\
     &low her to go to work.\\
     \hline
    \end{tabular}
    \caption{An example that illustrates the effectiveness of n-gram masking strategy in capturing distant context. \textcolor{blue}{Correct token}$/$\textcolor{red}{incorrect token} is marked in \textcolor{blue}{blue}$/$\textcolor{red}{red}.}
    \vspace{-1em}
    \label{label leakage}
\end{table}
\begin{table}[t]
\fontsize{8pt}{11pt}\selectfont
\centering
\begin{tabular}{|l|l|}
\hline
Input &\begin{CJK*}{UTF8}{gbsn}…去外国可以认识很多的人，就可以\textcolor{red}{借(bor-}\end{CJK*}\\
&\begin{CJK*}{UTF8}{gbsn}\textcolor{red}{row)}\textcolor{red}{少(less)}。
     \end{CJK*}\\
     \hline
     BERT &\begin{CJK*}{UTF8}{gbsn}
     …去外国可以认识很多的人，就可以\textcolor{red}{借(bor-}\end{CJK*}\\
     &\begin{CJK*}{UTF8}{gbsn}\textcolor{red}{row)}\textcolor{blue}{绍(mediate)}。
     \end{CJK*}\\
     \hline
     NM-BERT &\begin{CJK*}{UTF8}{gbsn}
     …去外国可以认识很多的人，就可以\textcolor{blue}{介绍(in-}\end{CJK*}\\
     &\begin{CJK*}{UTF8}{gbsn}
     \textcolor{blue}{troduce)}。
     \end{CJK*}\\
     \hline
     Trans & (Then) they can make a lot of friends in foreign  \\
     &countries and introduce (themselves to foreign-\\
     &ers)\\
     \hline
\end{tabular}
\caption{Examples that illustrate the effectiveness of n-gram masking strategy in blocking the error on adjacent positions.}
\vspace{-1em}
\label{error disturbance corrected}
\end{table}

\subsubsection{Case Analysis}

Table \ref{label leakage} shows an example where the correction depends on context in long distance. In this example, if given only the adjacent context \begin{CJK*}{UTF8}{gbsn}"让......去上班" \end{CJK*}(allow somebody to go to work), then either "allow him " or "allow her" is reasonable. But with the limitation of a more distant context that "the husband is selfish and asks his wife to stay at home", apparently the person who is not allowed to go to work is the wife and \begin{CJK*}{UTF8}{gbsn}"她"\end{CJK*}(her) is the correct character. ReaLiSe \citep{xu2021read} fails to take context in longer distance into account and remains the erroneous character\begin{CJK*}{UTF8}{gbsn}"他"\end{CJK*}(him) while our proposed method fixed this problem. 

Table \ref{error disturbance corrected} presents an example in which correction is disturbed by the adjacent error. As analyzed in table \ref{error disturbance}, BERT fails to correct \begin{CJK*}{UTF8}{gbsn}“借”(borrow)\end{CJK*} into \begin{CJK*}{UTF8}{gbsn}“介”\end{CJK*}(mediate) because of the error \begin{CJK*}{UTF8}{gbsn}“少”\end{CJK*}(less); while the proposed n-gram masked BERT corrected it successfully without being confused by the incorrect character. Note that this example is defined as consecutive errors in DCN\citep{wang2021dynamic}. DCN developed a complicated auto-regressive Dynamic Connected Scorer(DCScorer) to remedy this problem while this paper just utilizes a single transformer encoder layer and remains the non-autoregressive way of decoding which is more parameter efficient and faster. More cases with analysis are shown in Appendix \ref{appendix:cases}

\subsection{Conclusion}

This paper proposes a novel n-gram masking strategy to mitigate label leakage and error disturbance problem in Chinese Spell Checking(CSC). It further introduces a simple but effective dot-product gating mechanism to incorporate phonological and morphological knowledge into the CSC model. Experimental results show that the n-gram masking strategy improves all the base models on nearly all SIGHAN benchmarks. A simple n-gram masked BERT can even outperform sophisticated baselines such as ReaLiSe. Moreover, dot-product gating achieves better performance with more practical usage than previous state-of-the-art fusion strategies.

\section*{Limitations}
While n-gram masking strategy is proposed to block label on a correct position or noise on an erroneous position, it is not absolutely capable to achcieve this goal for 2 intrinsic designs in the transformer enoder layer.

The first one is residual connection denoted as Add\&Norm layer
in figure \ref{model}. As this figure shows, Add\&Norm takes the embedding output of the original sentence, i.e.$E_{out}$ as input to perform addition and layernorm, which enables the model to have a chance to "see" the label or error in the original sentence. The other model design that causes the same issue is the multi-layer structure. Even information of the current and adjacent tokens are masked in the n-gram masking layer, the semantic processing of the current position can still indirectly involve label or error leaked by a token in the next layer. That is, if the n-gram masking window of that token does not include the current position and its neighbors, that token may contain information from these positions.

An ideal solution would be employing $\rm{[mask]}$ token from the very beginning, take left-bigram masking as an example, if the input sentence is : 
\begin{table}[ht]
\centering
\begin{tabular}{c}
\begin{CJK*}{UTF8}{gbsn}习惯你的生活都很好。\end{CJK*}\\
(Supposed to mean: Hope you live a good life.)
\end{tabular}
\end{table}
\vspace{-1em}
then inputting the following masked sequences will block the information of the masked positions more thoroughly because the characters on these positions are invisible to every token in the sequence :
\begin{table}[ht]
\centering
\begin{tabular}{l|c}
\multicolumn{2}{c}{$\rm{[mask]}$ \begin{CJK*}{UTF8}{gbsn}惯你的生活都很好。\end{CJK*}}\\
\multicolumn{2}{c}{$\rm{[mask]}$$\rm{[mask]}$\begin{CJK*}{UTF8}{gbsn}你的生活都很好。\end{CJK*}}\\
\multicolumn{2}{c}{\begin{CJK*}{UTF8}{gbsn}习\end{CJK*}\begin{CJK*}{UTF8}{gbsn}$\rm{[mask]}$$\rm{[mask]}$的生活都很好。\end{CJK*}}\\
\multicolumn{2}{c}{\begin{CJK*}{UTF8}{gbsn}习惯$\rm{[mask]}$$\rm{[mask]}$生活都很好。\end{CJK*}}\\
\multicolumn{2}{l}{...}
\end{tabular}
\end{table}\\
However, this would increase the computational complexity dramatically and therefore impractical. The proposed method in this paper can be seen as a trade-off between performance and complexity, and future investigation is still necessary to completely solve the issues mentioned in this paper.

\bibliography{anthology,custom}
\bibliographystyle{acl_natbib}
\nocite{liu2021plome, zhang2021correcting, NIPS2017_3f5ee243, li2022past, li-2022-uchecker, hong2019faspell}

\clearpage
\appendix

\section{Appendix}

\subsection{Data Statistics}

\begin{table}[ht]
    \centering
    \small
    \renewcommand\arraystretch{1.2}
    \begin{tabular}{c|c|c|c}
    \hline
    Training Data & \#Line & Avg. Length & \#Errors\\
    \hline
    SIGHAN13 & 700 & 41.8 & 343\\
    SIGHAN14 & 3,437 & 49.6 & 5,122\\
    SIGHAN15 & 2,338 & 31.3 & 3,037\\
    WANG271K & 271,329 & 42.6 & 381,962\\
    \hline
    Total & 277,804 & 42.6 & 390,464\\
    \hline
    \hline
    Test Set & \#Line & Avg. Length & \#Errors\\
    \hline
    SIGHAN13 & 1,000 & 74.3 & 1,224\\
    SIGHAN14 & 1,062 & 50.0 & 771\\
    SIGHAN15 & 1,100 & 30.6 & 703\\
    \hline
    Total & 3,162 & 50.9 & 2,698\\
    \hline
    \end{tabular}
    \caption{Statistics of data.}
    \label{statics of data}
\end{table}
\label{appendix:label statistics}

\subsection{Experimental Setup}

This paper implements strict comparisons between the n-gram masked models and their corresponding baselines, where there is no difference in the data , implementation environment, way of model selection etc., between each baseline and its n-gram masked version.

ReaLiSe, PHMOSpell, BERT baseline, DGSpeller and their n-gram masked versions are initialized using the weights of  BERT-wwm while DCN is initialized by RoBERTa-wwm.\footnote{\url{https://github.com/ymcui/Chinese-BERT-wwm}} In the implementations of all the models mentioned above, an AdamW optimizer \citep{ilya2019decoupled} with a learning rate of 5e-5 is utilized in training. All training data in table \ref{statics of data} are trained for 10 epochs with a batch size of 16 per GPU. 2 Tesla P40 with 24G memory are available, therefore total batch size is 32. Furthermore, model of the last training step is directly saved for evaluation, and all "de" corrections in SIGHAN13 are removed in the evaluation following the convention of ReaLiSe.

For SpellGCN, Chinese BERT released by Google Research\footnote{\url{https://github.com/google-research/bert}} is used for initialization. The training conforms to a 2-stage convention: in the first stage, all the training data presented in table \ref{statics of data} are trained for 10 epochs; in the second stage, the model of the last step in the first training stage are finetuned on the training set of SIGHAN13 and all SIGHAN training data respectively for additional 6 epochs. The model finetuned on SIGHAN13 is used for the evaluation of SIGHAN13 test set while the one finetuned on all SIGHAN training data is utilized in the evaluation of test sets of SIGHAN14 and SIGHAN15. Note that "de" corrections are not removed from the evaluation of SIGHAN13 following the same convention in the original paper. An AdamW optimizer with a learning rate of 5e-5 is adopted in the 2-stage training, where 1 Tesla P40 with 24G memory is used and the batch size is 32.

All the baseline models except PHMOSpell are implemented using their officially released code which are licensed by an Apache 2.0 or MIT license\footnote{\url{https://github.com/ACL2020SpellGCN/SpellGCN}}\footnote{\url{https://github.com/DaDaMrX/ReaLiSe}}\footnote{\url{https://github.com/destwang/dcn}}. For PHMOSpell, since the official release is not available, this paper reimplemented it based on ReaLiSe. Generally, sophiscated encoders like transformer or ResNet would lead to better performance than a simple embedding lookup table. Therefore, in order to assure the fairness of comparison between different fusion strategies, the embedding tables in the original paper of PHMOSpell are replaced with phonetic and graphic encoders defined in ReaLiSe. In other words, ReaLiSe, PHMOSpell and DGSPeller share the same phonological and morphological encoders in this paper. In the original paper of PHMOSpell, the weights for pinyin feature and glyph feature are 0.6,0.4 for SIGHAN13, and 0.8, 0.2 for SIGHAN14 and SIGHAN15, while in this paper these 2 groups of weights are used to train 2 models, and the better result out of 2 settings is reported for each sighan test set, i.e. 0.6, 0.4 for SIGHAN13 and SIGHAN14; 0.8,0.2 for SIGHAN15.

All the n-gram versions of the baseline models are implemented using exactly the same settings as their corresponding baselines on the same environnment except adopting an n-gram masking strategy. Experiments are conducted in unigram, bigram  and trigram settings, and the best results are reported.
\label{appendix: experimental setup}

\subsection{Cases}

Table \ref{label leakage 2} shows 3 examples whose corrections depend on  contexts in a long distance. In the first example, affected by the adjacent character \begin{CJK*}{UTF8}{gbsn}"心"\end{CJK*}, ReaLiSe \citep{xu2021read} has chosen \begin{CJK*}{UTF8}{gbsn}"情"\end{CJK*} as correction which often co-occurs with \begin{CJK*}{UTF8}{gbsn}"心"\end{CJK*} to form a common expression \begin{CJK*}{UTF8}{gbsn}"心情"\end{CJK*}(mood). However, \begin{CJK*}{UTF8}{gbsn}"善良"\end{CJK*}(kind-hearted) in a more distant context, which is an adjective to describe human's nature, indicates that \begin{CJK*}{UTF8}{gbsn}"地"\end{CJK*}(nature) is already the correct character, and our proposed method did not over-correct it. Similarily in the second example, \begin{CJK*}{UTF8}{gbsn}“杀戒”(forbiddance of killing) \end{CJK*} is a common expression, but with the context "we really like it and  want to buy it ", the correct character should be \begin{CJK*}{UTF8}{gbsn}“价”(price)\end{CJK*}. In the third example, \begin{CJK*}{UTF8}{gbsn}“支”\end{CJK*} is a quantifier specific to something without life. The correct character should be \begin{CJK*}{UTF8}{gbsn}“只”\end{CJK*} because it can be inferred from a more distant context that the speaker is talking about the dog, which is a living thing.

\begin{table}[ht]
    \fontsize{8pt}{12pt}\selectfont
    \centering
    \begin{tabular}{|l|l|}
     \hline
     Input & \begin{CJK*}{UTF8}{gbsn}
     …老师们很聪明，心\textcolor{blue}{地(nature)}也很善良。    
     \end{CJK*}\\
     \hline
     BERT & \begin{CJK*}{UTF8}{gbsn}
     …老师们很聪明，心\textcolor{red}{情(mood)}也很善良。    
     \end{CJK*}\\
     \hline
     NM-BERT & \begin{CJK*}{UTF8}{gbsn}
     …老师们很聪明，心\textcolor{blue}{地(nature)}也很善良。
     \end{CJK*}\\
     \hline
     Trans & ... the teachers are smart and good-natured.\\
     \hline
     Input & \begin{CJK*}{UTF8}{gbsn}
     …我们很喜欢很想买，可是对我们来说一定\end{CJK*}\\
     &\begin{CJK*}{UTF8}{gbsn}要杀\textcolor{red}{戒(forbiddance)}一下。\end{CJK*}\\
     \hline
     BERT & \begin{CJK*}{UTF8}{gbsn}
     …我们很喜欢很想买，可是对我们来说一定\end{CJK*}\\
     &\begin{CJK*}{UTF8}{gbsn}要杀\textcolor{red}{戒(forbiddance)}一下。\end{CJK*}\\
     \hline
      NM-BERT & \begin{CJK*}{UTF8}{gbsn}
     …我们很喜欢很想买，可是对我们来说一定\end{CJK*}\\
     &\begin{CJK*}{UTF8}{gbsn}要杀\textcolor{blue}{价(price)}一下。\end{CJK*}\\
     \hline
     Trans & We really like it and want to buy it, but we still
     \\
     &need to bargain.\\
     \hline
     Input & \begin{CJK*}{UTF8}{gbsn}
     我也会带我的小狗去，你大概知道是哪一\textcolor{red}{支}\end{CJK*}\\
     &\begin{CJK*}{UTF8}{gbsn}\textcolor{red}{(quantifier)}，它的眼睛蓝蓝的，圆圆的。\end{CJK*}\\
     \hline
     BERT & \begin{CJK*}{UTF8}{gbsn}
     我也会带我的小狗去，你大概知道是哪一\textcolor{red}{支}\end{CJK*}\\
     &\begin{CJK*}{UTF8}{gbsn}\textcolor{red}{(quantifier)}，它的眼睛蓝蓝的，圆圆的。\end{CJK*}\\
     \hline
     NM-BERT & \begin{CJK*}{UTF8}{gbsn}
     我也会带我的小狗去，你大概知道是哪一\textcolor{blue}{只}\end{CJK*}\\
     &\begin{CJK*}{UTF8}{gbsn}\textcolor{blue}{(quantifier)}，它的眼睛蓝蓝的，圆圆的。\end{CJK*}\\
     \hline
     Trans & I will also take my dog. You may know which \\
     & one it is. It has blue and round eyes.
     \\
     \hline
    \end{tabular}
    \caption{Examples that illustrate the effectiveness of n-gram masking strategy in capturing distant context. \textcolor{blue}{Correct token}$/$\textcolor{red}{incorrect token} is marked in \textcolor{blue}{blue}$/$\textcolor{red}{red}.}
    \label{label leakage 2}
\end{table}

\begin{table}[ht]
\fontsize{8pt}{11pt}\selectfont
\centering
\begin{tabular}{|l|l|}
     \hline
     Input &\begin{CJK*}{UTF8}{gbsn}我的好朋友，你好！\textcolor{red}{习惯(habit)}你的生活都\end{CJK*}\\
     &\begin{CJK*}{UTF8}{gbsn}很好。\end{CJK*}\\
     \hline
     ReaLiSe &\begin{CJK*}{UTF8}{gbsn}我的好朋友，你好！\textcolor{blue}{希(hope)}\textcolor{red}{惯(habit)}你的\end{CJK*}\\
     &\begin{CJK*}{UTF8}{gbsn}生活都很好\end{CJK*}\\
     \hline
     NM-ReaLiSe&\begin{CJK*}{UTF8}{gbsn}我的好朋友，你好！\textcolor{blue}{希}\textcolor{blue}{望(hope)}你的生活都\end{CJK*}\\
     &\begin{CJK*}{UTF8}{gbsn}很好。\end{CJK*}\\
     \hline
     Trans & My good friend, hello! Hope you live a good\\
     &life.\\
     \hline
     Input & \begin{CJK*}{UTF8}{gbsn}她的脸\textcolor{red}{百百(hundred)}的、\textcolor{red}{园园(garden)}的，\end{CJK*}\\
     &\begin{CJK*}{UTF8}{gbsn}头发黑黑的，眼睛大大的。\end{CJK*}\\
     \hline
     BERT & \begin{CJK*}{UTF8}{gbsn}她的脸\textcolor{blue}{白白(white)}的、\textcolor{blue}{圆(round)}\textcolor{red}{园(garden)}\end{CJK*}\\
     &\begin{CJK*}{UTF8}{gbsn}的，头发黑黑的，眼睛大大的。\end{CJK*}\\
     \hline
     NM-BERT & \begin{CJK*}{UTF8}{gbsn}她的脸\textcolor{blue}{白白(white)}的、\textcolor{blue}{圆圆(round)}的，头\end{CJK*}\\
     &\begin{CJK*}{UTF8}{gbsn}发黑黑的，眼睛大大的。\end{CJK*}\\
     \hline 
     Trans& She has a white and round face with black \\
     &hair and big eyes.\\
     \hline
\end{tabular}
\caption{Examples that illustrate the robustness of n-gram masking to adjacent errors.}
\label{error disturbance corrected 2}
\end{table}
\label{appendix:cases}

\end{document}